\documentclass[conference]{IEEEtran}

\usepackage[cmex10]{amsmath}
\usepackage{amsthm}
\usepackage{amssymb}
\usepackage{mathrsfs}
\usepackage{graphicx}
\usepackage{float}
\usepackage{array}
\usepackage{epstopdf}
\usepackage{multirow}
\usepackage{cite}

\begin{document}

\title{F\MakeLowercase{inger}-GAN:
G\MakeLowercase{enerating}
R\MakeLowercase{ealistic}
F\MakeLowercase{ingerprint} I\MakeLowercase{mages}
U\MakeLowercase{sing}
C\MakeLowercase{onnectivity}
I\MakeLowercase{mposed}
GAN}

\author{Shervin Minaee$^*$, AmirAli Abdolrashidi$^{\dagger}$  \\
$^*$New York University
\\ $^{\dagger}$University of California, Riverside\\ \\
}

\maketitle

\begin{abstract}
Generating realistic biometric images has been an interesting and, at the same time, challenging problem. 
Classical statistical models fail to generate realistic-looking fingerprint images, as they are not powerful enough to capture the complicated texture representation in fingerprint images.
In this work, we present a machine learning framework based on generative adversarial networks (GAN), which is able to generate fingerprint images sampled from a prior distribution (learned from a set of training images).
We also add a suitable regularization term to the loss function, to impose the connectivity of generated fingerprint images. 
This is highly desirable for fingerprints, as the lines in each finger are usually connected. 
We apply this framework to two popular fingerprint databases, and generate images which look very realistic, and similar to the samples in those databases.
Through experimental results, we show that the generated fingerprint images have a good diversity, and are able to capture different parts of the prior distribution.
We also evaluate the Frechet Inception distance (FID) of our proposed model, and show that our model is able to achieve good quantitative performance in terms of this score.
\end{abstract}

\IEEEpeerreviewmaketitle

\section{Introduction}
\label{sec:Intro}
Fingerprint is arguably the most popular biometric used for identification and security. Simple to collect and rich in features, it provides decent benefits for an efficient recognition system. 
Although it is not as protected as features such as iris pattern \cite{daugman_howirisworks}, fingerprint recognition is still among the most researched topics in biometrics along with iris \cite{iris_intro} and face \cite{face_intro}, and is popularly used in forensics, airports and smartphones (such as in Iphone). All biometrics have unique features and patterns. Therefore, generating a synthetic fingerprint is a challenging effort. Our work seeks to accomplish this and measure the level of authenticity against real fingerprints.

Many approaches have been proposed for fingerprint recognition, which are based on matching minutiae features, such as ridge endings and bifurcation \cite{finger_old1, finger_old2}.
There are also several approaches based on hand-crafted features extracted from fingerprint images \cite{handcraft1, handcraft2}.
Also, with deep neural networks increasing in popularity in recent years \cite{cnn1,cnn2}, end-to-end models have been also used for biometrics recognition  \cite{finger_new1, finger_new2, irisdeep1, irisdeep2, palmdeep}.


\begin{figure}[h]
\begin{center}
   \includegraphics[width=0.9\linewidth]{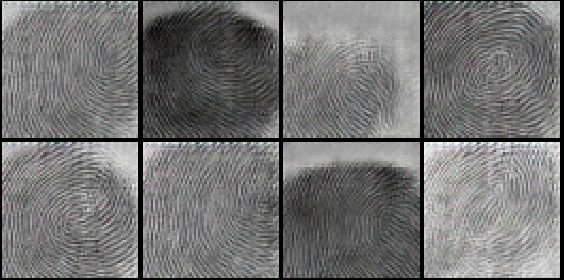}
\end{center}
   \caption{Eight sample generated fingerprint images using the proposed framework. These images are generated with a model trained on PolyU fingerprint database.}
\label{fig:gan_model}
\end{figure}

In this work, we focus on a different problem from fingerprint recognition, which is learning to generate fingerprint images.
Our work is mainly based on generative adversarial networks (GAN) \cite{gan}, which provides a powerful framework for learning to generate  samples from a given distribution. GAN is comprised of a \textit{generative network}, which learns a training dataset's distribution and can generate new data (sampled from the same distribution), and a \textit{discriminative network}, which tries to distinguish real samples from the ones generated with the generator model. 
On a high level, the generator and discriminator models are trained simultaneously, using a minimax type optimization. 
Since GAN's invention, it has been used for various computer vision and pattern generation tasks, such as image super-resolution, image to image translation, image generation based on text description, prediction of pedestrian walking patterns for autonomous vehicles, and iris generation \cite{sr-gan, pix2pix, gan_text, gan_walk, irisgan}.

In this work, we propose a synthetic fingerprint generation framework based on deep convolutional generative adversarial networks (DC-GAN), which is able to generate realistic fingerprint images, which are hard to be distinguished from real ones. 
By far, to the best of our knowledge, no other work in literature has tackled fingerprint generation using GANs.
We add  a suitable term to the GAN loss function, imposing the generated fingerprint images to have connected lines. This is achieved by adding the total variation \cite{TV} of the generated images to the loss function. 
Figure \ref{fig:gan_model} provide 8 sample images generated by our proposed model.

The structure of the rest of this paper is as follows.
Section \ref{sec:Framework} provides the details of our proposed framework, and the model architectures for both the generator, and discriminator networks.
In Section \ref{sec:Evaluation}, we provide the experimental studies, and present the generated fingerprint images with our proposed framework. We show the generated fingerprint images over different epochs (for the sample input noise) to see how the generated images evolve over time, and finally the paper is concluded in Section \ref{sec:Conclusion}.

\section{The Proposed Framework}
\label{sec:Framework}
In this work we propose a generative model for fingerprint images based on deep convolutional generative adversarial networks, with a suitable regularization which suits fingerprint generation.
Let us first give more technical details on GAN, and then we dive into our proposed model.

\subsection{Generative Adversarial Network}
\begin{figure}[h]
\begin{center}
   \includegraphics[page=1,width=0.98\linewidth]{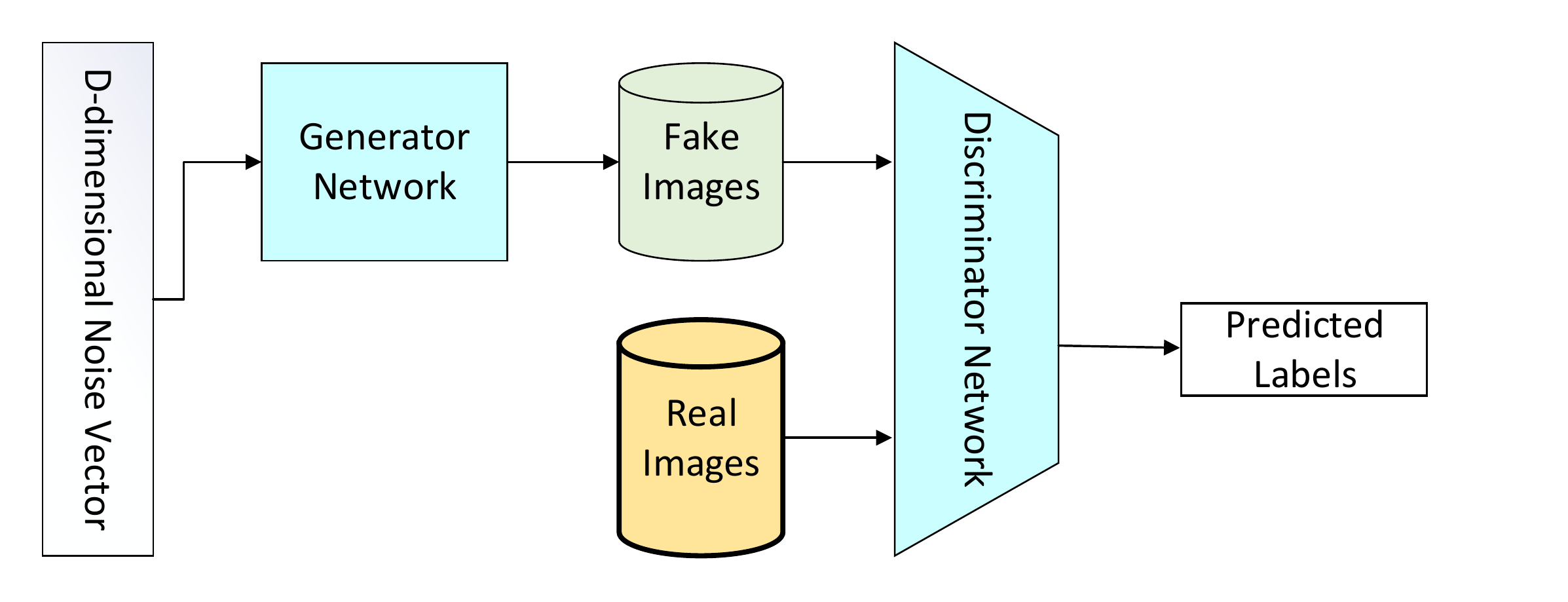}
\end{center}
   \caption{The general architecture of generative adversarial networks}
\label{fig:gen_arch}
\end{figure}

To provide more details about how GANs work, the generator network in GANs learns a mapping from noise $z$ (with a prior distribution) to a target distribution $y$, $G= z \rightarrow y$.
The generator model, $G$, tries to generate samples which look similar to the real samples (provided during training), while the discriminator network, $D$, tries to distinguish the samples generated by the generator models from the real ones.
The general architecture of a GAN model is shown in Figure \ref{fig:gen_arch}.

The generator and discriminator can be trained jointly, by minimizing a loss function.
The loss function of GAN can be written as:
\begin{equation}
\begin{aligned}
& \mathcal{L}_{GAN}=  \mathbb{E}_{x \sim p_{data}(x)}[\text{log} D(x)]+ \mathbb{E}_{z \sim p_z(z)}[\text{log}(1-D(G(z)))]
\end{aligned}
\end{equation}
In this sense we can think of GAN, as a minimax game between $D$ and $G$, where $D$ is trying to minimize its classification error in distinguishing the fake samples from the real ones (maximize $\mathcal{L}_{GAN}$), and $G$ is trying to maximize the discriminator network's error (minimize $\mathcal{L}_{GAN}$).
After training this model, the trained generator model would be: 
\begin{equation}
\begin{aligned}
G^*= \text{arg} \ \text{min}_G \text{max}_D \ \mathcal{L}_{GAN}
\end{aligned}
\end{equation}
In practice, $\mathcal{L}_{GAN}$ may not provide enough gradient for $G$ to get trained well, specially at the early steps of the training where the discriminator can easily detect fake samples from the real ones. 
One solution is to let the generator to maximize $\mathbb{E}_{z \sim p_z(z)}[\text{log}(D(G(z)))]$ (instead of minimizing $\mathbb{E}_{z \sim p_z(z)}[\text{log}(1-D(G(z)))]$).


As discussed briefly in the introduction, there have been several works improving on GAN on one or all directions.
Since invention of GAN, there have been several works trying to improve/modify GAN in different aspects.
To name a few promising works, in \cite{dc-gan}, Radford and colleagues proposed a convolutional GAN model for generating images, which generates visually higher quality images than fully connected networks.
In \cite{con-gan}, Mirza et al. proposed a conditional GAN model, which can generate images conditioned on class labels. 
This enables one to generate samples with a specified label (such as specific digit when trained on MNIST dataset).
The original paper on GAN uses KL divergence to measure the distance between distributions, and a problem can happen if these distribution have non-overlapping support in which KL divergence is not a good representative of their distance. 
The work in \cite{was-gan} addresses this issue, by proposing a new loss function based on Wasserstein distance (also known as earth mover's distance).
In \cite{cycle-gan}, Zhu proposed an image to image translation model based on a cycle-consistent GAN model, which learns to map a given image distribution into a target domain (e.g. horse to zebra, day to night images, summer to winter images).
In \cite{gan_text}, Zhang et al proposed a GAN-based high-resolution image generation network using a text description of the image.
The idea of adversarial training has been also applied to auto-encoder framework, to provide an unsupervised feature learning approach, e.g. in \cite{adv_ae1, adv_ae2},  where the adversarial module is trying to distinguish samples for which the latent representation is coming from a prior distribution from other samples.

\subsection{TV-Regularized GAN}
We use a deep convolutional GAN (DC-GAN) as the main core of our fingerprint generation framework.
The fingerprint images consist of many lines, each forming a connected component, and help the model to generate fingerprint images with their lines being connected (instead of having broken lines).
Although the GAN model should learn this line connectivity from the training data, we found out imposing a regularization term which promotes the connectivity of the generated fingerprint images can help the generator learn this more easily.

There are several ways to impose connectivity within images, such as total variation (TV), group sparsity \cite{TV, group-sparse}. 
Total variation penalizes the solutions (in this case, the generated images) with large variations among neighboring pixels, leading to more connected and smoother solutions \cite{TV1, TV2, TV3}.
We chose to use total variation to impose the connectivity of generated images here, as its gradient update is less complex during back-propagation. 
Total variation of a differentiable function $f$, defined on an interval $[a,b] \subset R$, has the following expression if $f'$ is Riemann-integrable:
\begin{equation}
\begin{aligned}
V_a^b= \int_{a}^{b} \| f' (x) \| dx
\end{aligned}
\end{equation}

Total variation of 1D discrete signals $y=[y_1, y_2,  ..., y_N]$ is straightforward, and can be defined as:
\begin{equation}
\begin{aligned}
TV(y)=  \sum_{n=1}^{N-1} |y_{n+1}-y_{n}|
\end{aligned}
\end{equation}

For 2D signals, $Y= [y_{i,j}]$, we can use the isotropic or the anisotropic version of 2D total variation \cite{TV}. 
To make our optimization problem simpler, we have used the anisotropic version of total variation, which is defined as the sum of horizontal and vertical gradients at each pixel:
\begin{equation}
\begin{aligned}
TV(Y)=  \sum_{i,j} |y_{i+1,j}-y_{i,j}|+|y_{i,j+1}-y_{i,j}|
\end{aligned}
\end{equation}

We can add the total variation of the generated image to the loss function to promote the connectivity of the generated fingerprint images.
Adding this term to our framework, the new loss function for our model would be defined as:
\begin{equation}
\begin{aligned}
&  \mathcal{L}_{GAN-TV}=  \mathbb{E}_{x \sim p_{data}(x)}[\text{log} D(x)]+ \\ & \mathbb{E}_{z \sim p_z(z)}[\text{log}(1-D(G(z)))] + \lambda \ \text{TV}(G(z))
\end{aligned}
\end{equation}
\\We can then train both the generator and discriminator by stochastic gradient descent on mini-batches of training images.

\subsection{Network Architecture}
\begin{figure}[h]
\begin{center}
   \includegraphics[page=2,width=1\linewidth]{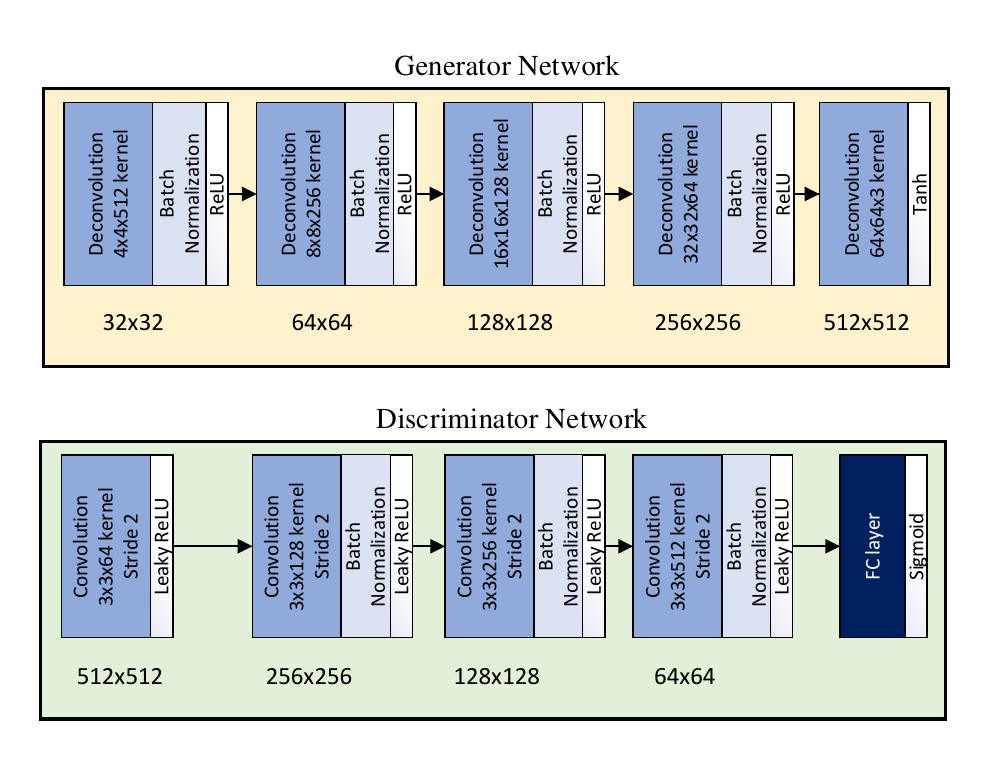}
\end{center}
   \caption{The architecture of the proposed generator (top) and discriminator model (bottom) by our work.}
\label{fig:our_model}
\end{figure}

We will now discuss the architectures used for our discriminator and generator networks.
As discussed above, both the discriminator and generator networks in our model are based on convolutional neural networks.
The discriminator model consist of 4 convolutional layers followed by batch normalization and leaky ReLU as non-linearity, and a fully connected layer at the end.
The generator model contains 5 fractionally-strided convolutional layers (aka deconvolution in literature), followed by batch-normalization and non-linearity.
The architecture of our generator and discriminator networks are shown in Figure \ref{fig:our_model}.

\section{Experimental Results}
\label{sec:Evaluation}
Before presenting the generated images by the proposed model, let us first discuss the hyper-parameters used in our training procedure.
We train the proposed model for 120 epochs on a Nvidia Tesla GPU.
The batch size is set to 40, and ADAM optimizer is used to optimize the loss function, with a learning rate of 0.0002.
The input noise to the generator network, is 100-dimensional Gaussian distribution with zero mean, unit variance.
We present the details of the datasets used for our work in the next section, and then present experimental results.

\subsection{Databases}
In this section, we provide a quick overview of two popular fingerprint databases used in our work, FVC-2006 fingerprint dataset \cite{FVC2006}, and PolyU fingerprint dataset \cite{polyU}.

\begin{figure}[ht]
\begin{center}
   \includegraphics[width=0.9\linewidth]{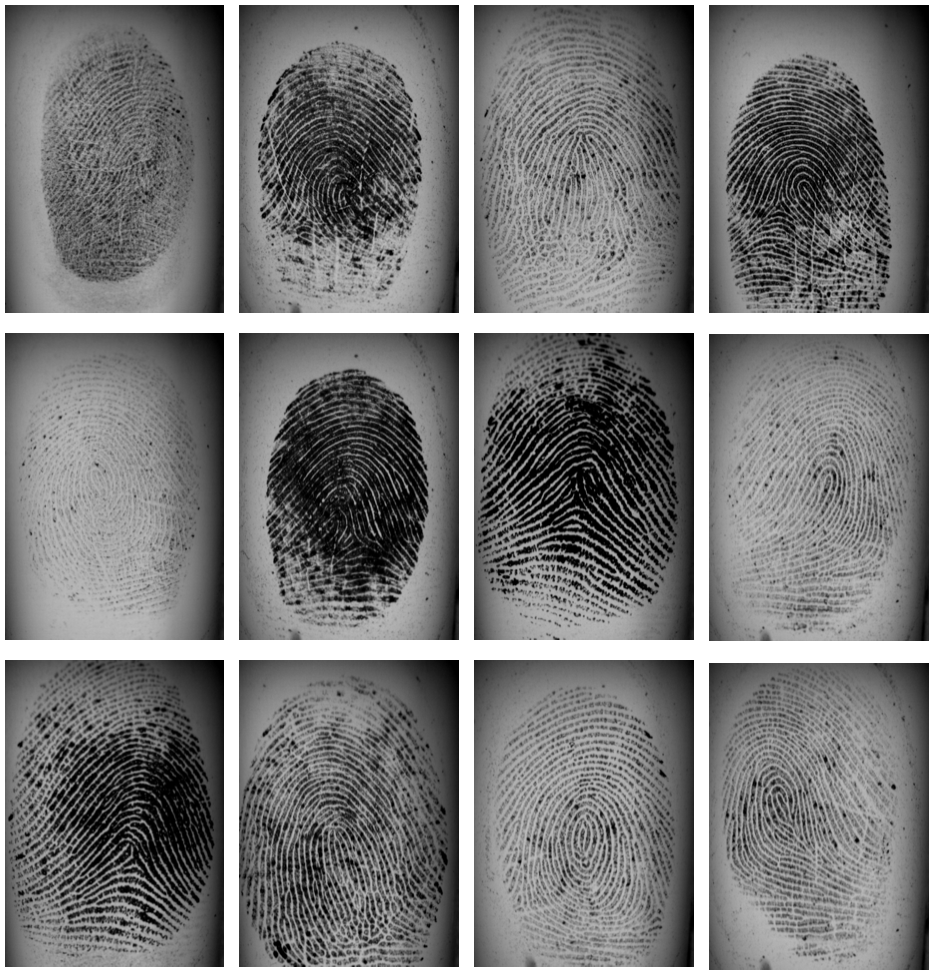}
\end{center}
   \caption{12 sample images from FVC-2006  fingerprint database.}
\label{fig:fvc2006}
\end{figure}

\textbf{FVC 2006 Fingerprint Database}:
FVC has provided four distinct fingerprint databases, DB1, DB2, DB3 and DB4. 
Each database is 150 fingers wide and 12 samples per finger in depth (i.e., it consists of 1800 fingerprint images). 
Each database is partitioned in two disjoint subsets A and B, which contain the first 140 fingers (1680 images) and  the last 10 fingers (120 images) 
respectively. 
The image size and resolution vary depending on the database. 
For further details about this dataset, we refer the readers to \cite{FVC_site}.
In our work, we have used the DB2-A dataset, which is taken by an ``Optical Sensor'', and has a resolution of 400x560.
Twelve sample images from this database are shown in Figure \ref{fig:fvc2006}.
It is worth mentioning that we first extract the central crop (400x400) of these images, and then resize them to 64x64 before feeding them to the TV-regularized GAN model. We also perform some data augmentation to increase the number of images during training.

\textbf{PolyU Fingerprint Databases}:
PolyU (high-resolution) fingerprint database is provided by the Biometrics Research Centre of the Hong Kong Polytechnic University.
They provide two high-resolution fingerprint image databases (denoted as DB-I and DB-II) \cite{Poly_site}.
DB-I consists of 1600 images of 320x240 resolution, and DB-II provides 1480 images of 640x480.
In this work, we used the DB-II database.
We first extract the central crop (480x480) of these images, and then resize them to 64x64 before feeding them to our TV-regularized GAN model.

\begin{figure}[h]
\begin{center}
   \includegraphics[width=0.8\linewidth]{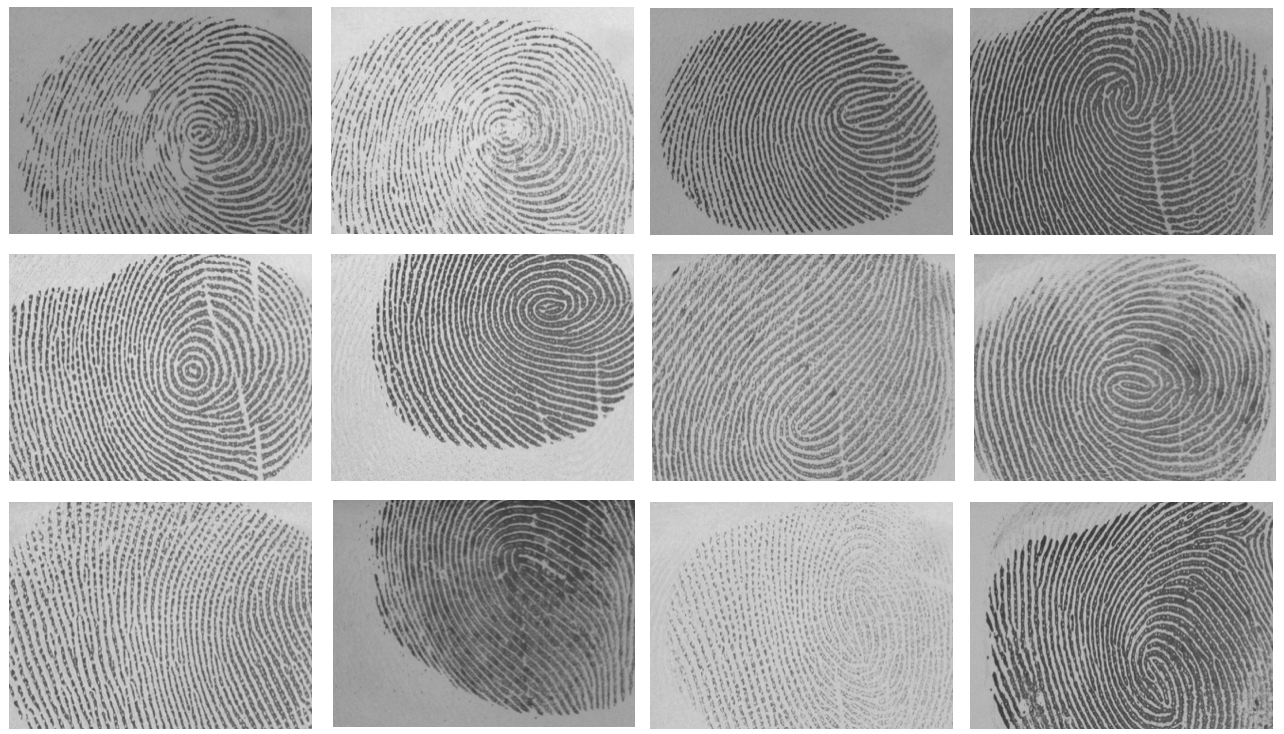}
\end{center}
   \caption{12 sample images from PolyU fingerprint database.}
\label{fig:polyU12}
\end{figure}

Twelve sample images from this database are shown in Figure \ref{fig:polyU12}.

\subsection{Experimental Analysis and Generated Fingerprint Images}
We will now present the experimental results of the proposed fingerprint generator model trained on the above databases.
We trained one model per dataset in our experiments, with the model parameters discussed above.
After training these models, we can generate different images by sampling latent representation from our prior Gaussian distribution and feeding them to the generator network of the trained models.

\begin{figure}[ht]
\begin{center}
    \begin{tabular}{cc}
    \rotatebox{0}{0th Epoch} &
    \rotatebox{0}{20th Epoch} \\
    \includegraphics[width=0.45\linewidth]{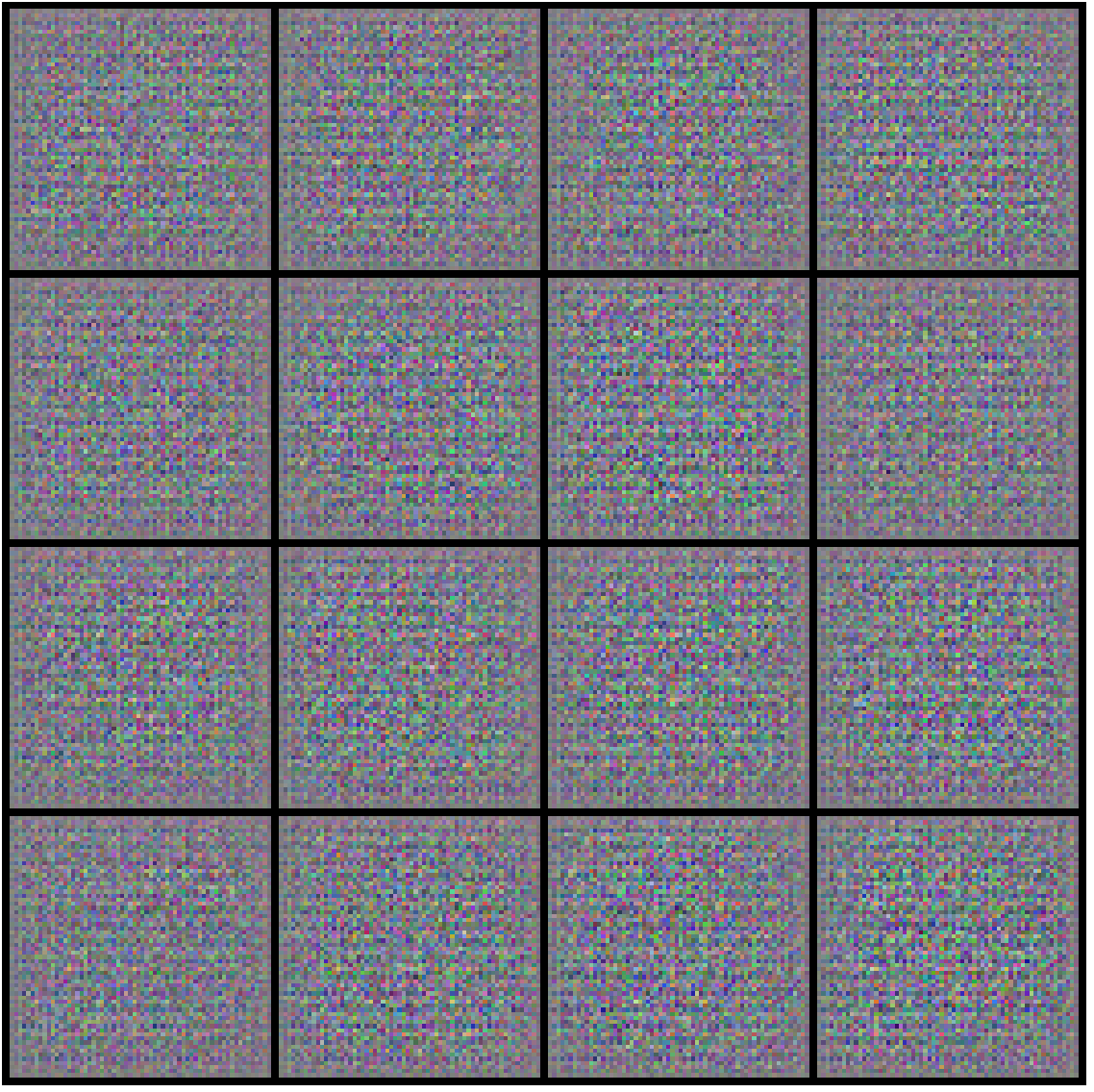}
    & \includegraphics[width=0.45\linewidth]{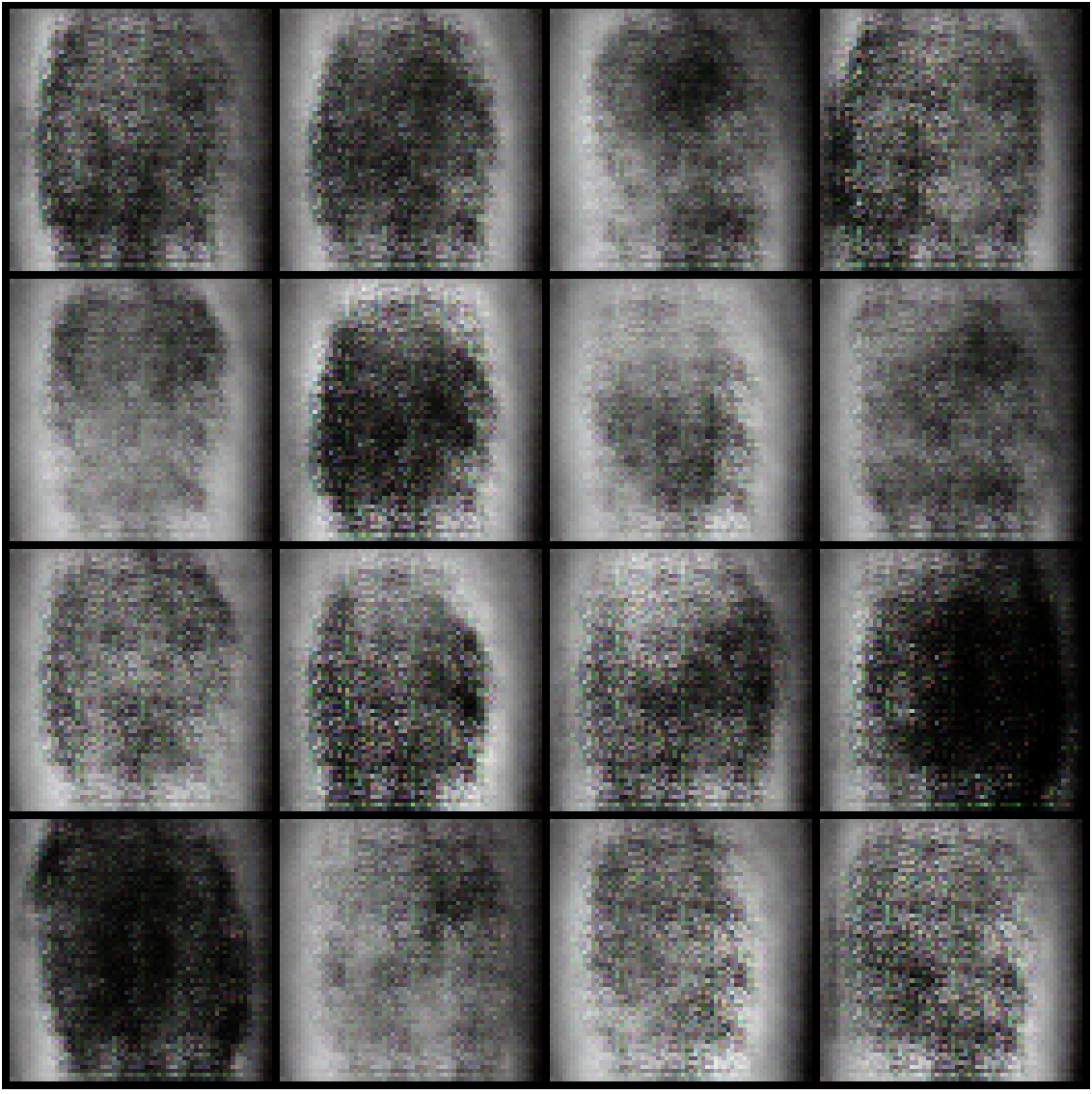}\\ 
    \rotatebox{0}{40th Epoch} &
    \rotatebox{0}{60th Epoch} \\
    \includegraphics[width=0.45\linewidth]{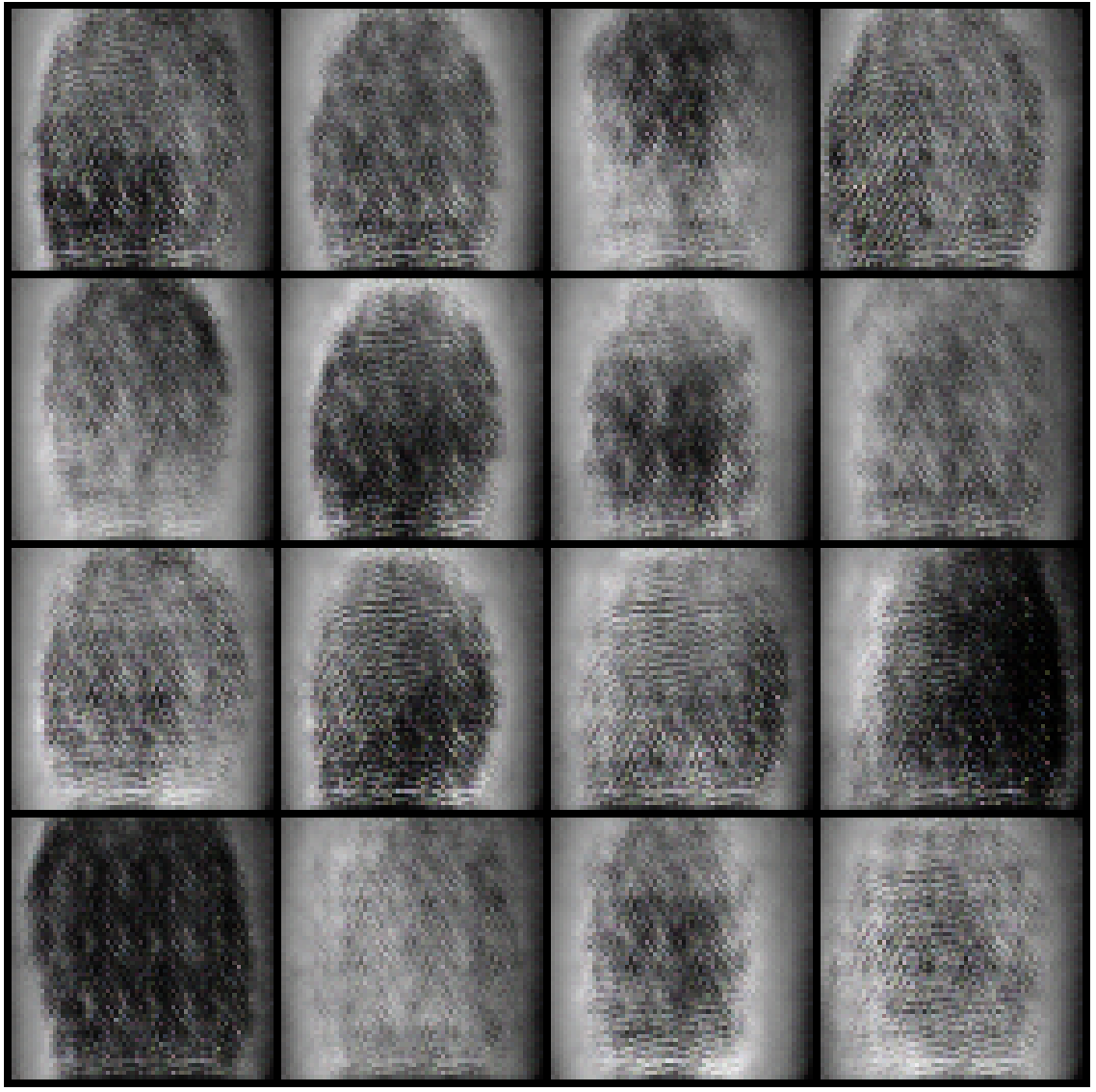}
    & \includegraphics[width=0.45\linewidth]{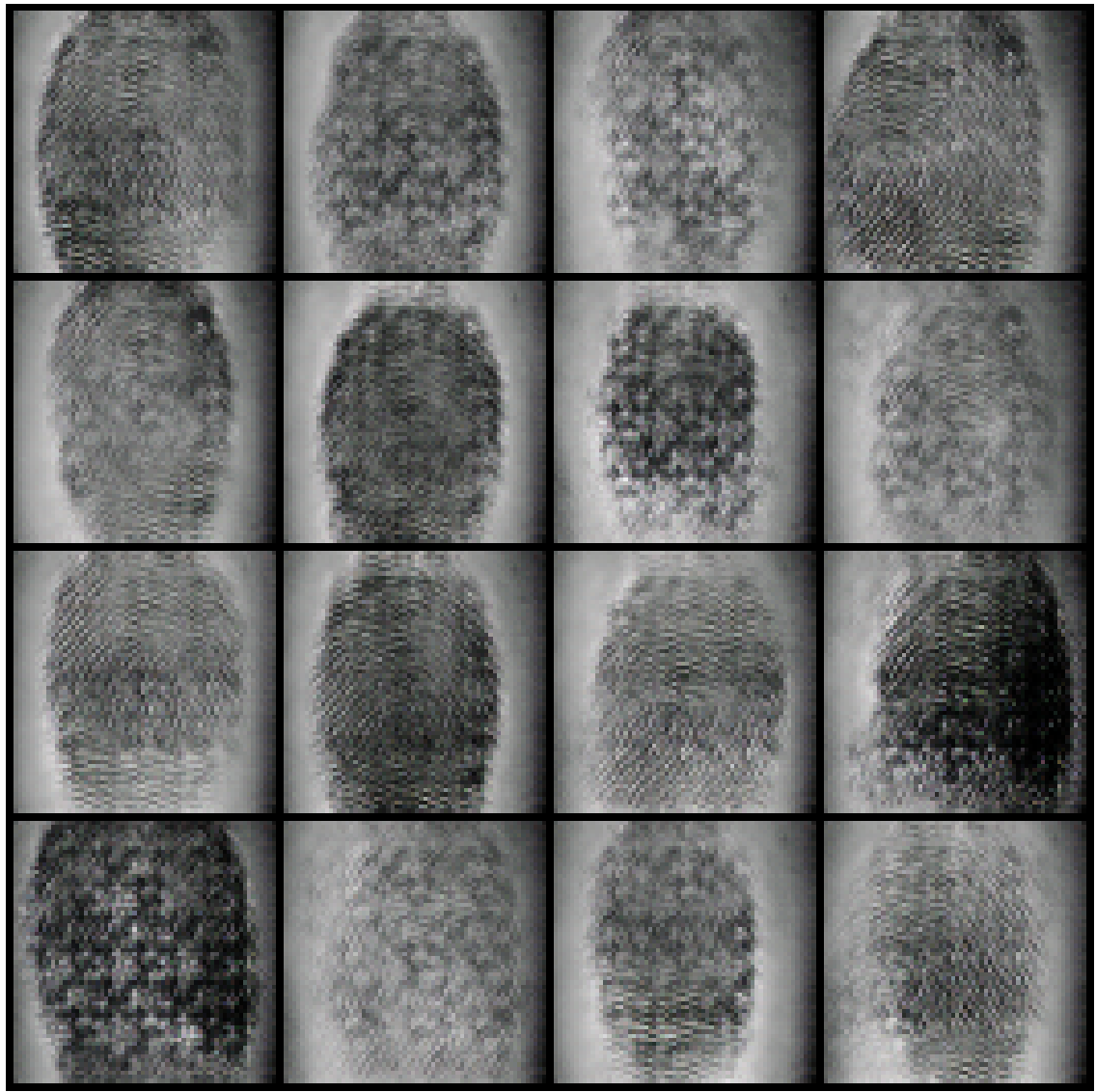}\\ 
    \rotatebox{0}{80th Epoch} &
    \rotatebox{0}{100th Epoch} \\
    \includegraphics[width=0.45\linewidth]{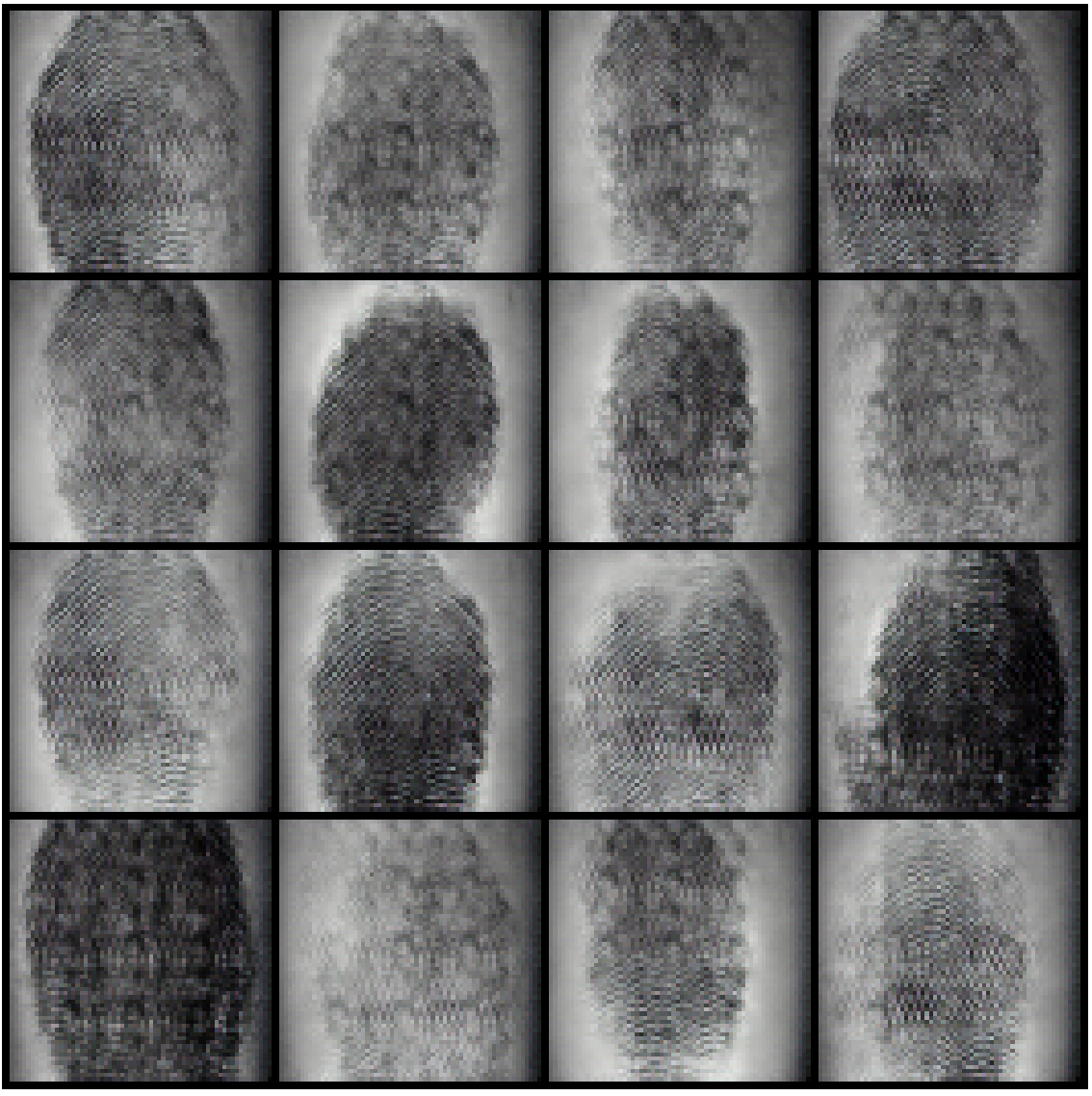}
    & \includegraphics[width=0.45\linewidth]{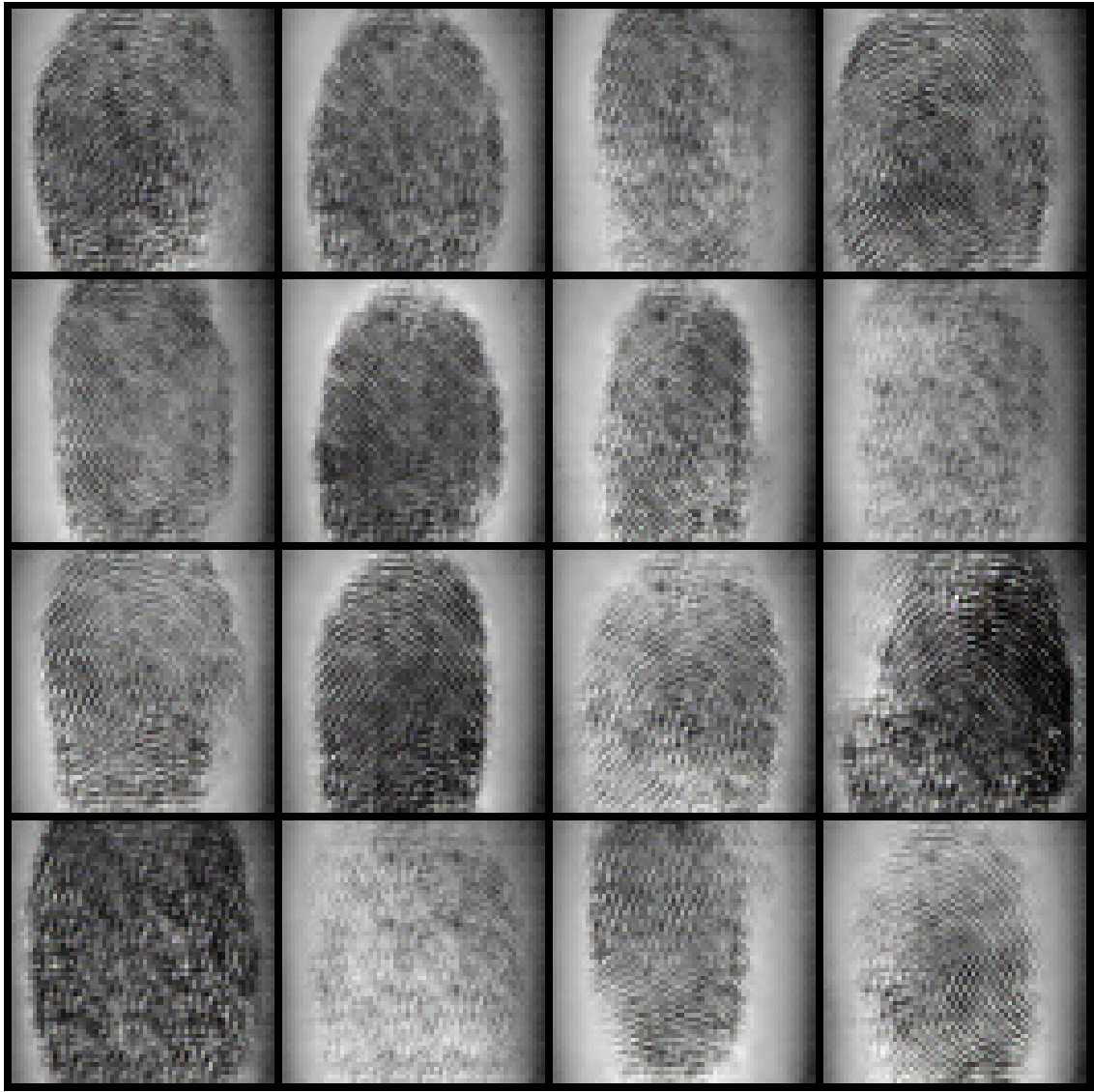}\\ 
    \end{tabular}
\end{center}
   \caption{16 generated fingerprint images using the same input noise, over different epochs.}
\label{fig:polyU16}
\end{figure}

In Figure \ref{fig:polyU16}, we present 16 generated fingerprint images over different epochs (0th, 20th, 40th, 60th, 80th and 100th), to see the amount of diversity among the generated images (in terms of the position of the width, height, and color of the fingerprints).
As it can be seen, there is a good amount of diversity in all the aforementioned aspects across the 16 generated images.

\begin{figure*}[t]
\begin{center}
   \includegraphics[width=0.99\linewidth, trim={0 1cm 0 0}]{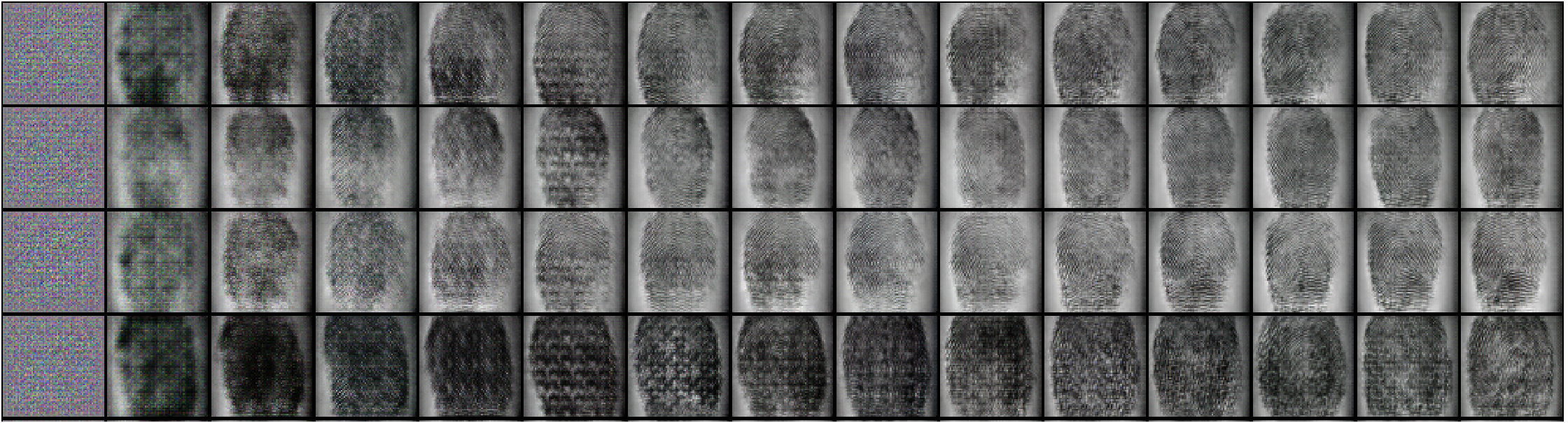}
\end{center}
   \caption{The generated fingerprint images for 4 input latent vectors, over 120 epochs (on every 10 epochs), using the trained model on FVC-2006 fingerprint database.}
\label{fig:fvc2006_generate}
\end{figure*}

\begin{figure*}[t]
\begin{center}
   \includegraphics[width=0.99\linewidth, trim={0 1cm 0 0}]{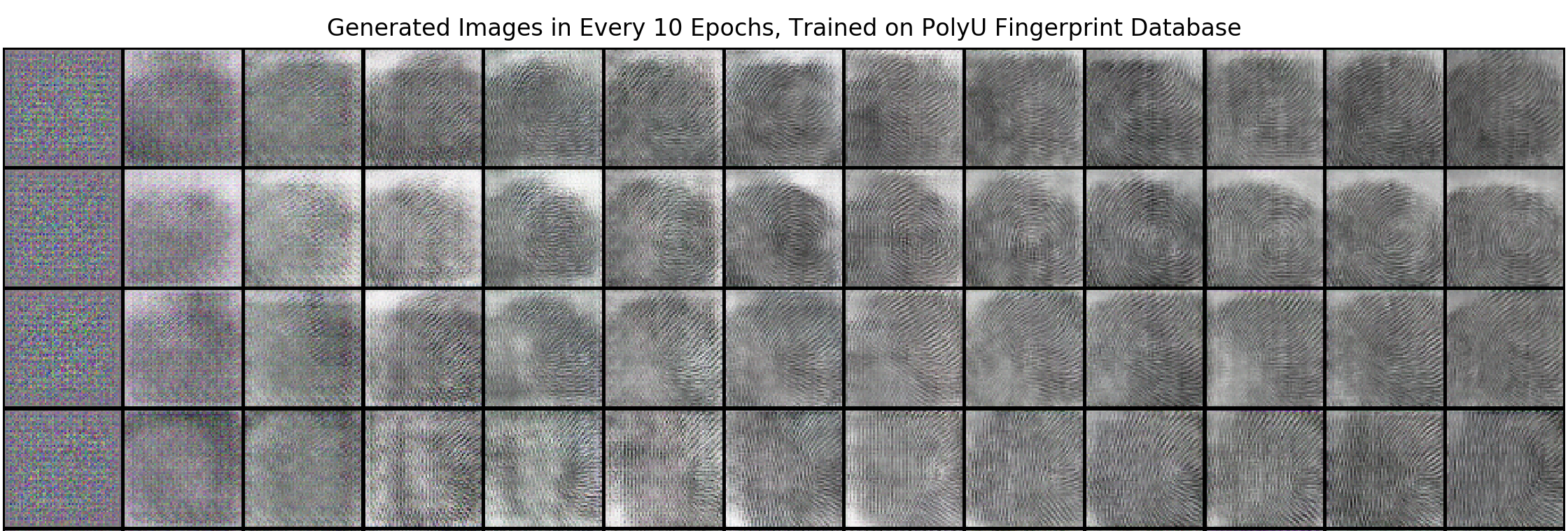}
\end{center}
   \caption{The generated fingerprint images for 4 input latent vectors, over 120 epochs (on every 10 epochs), using the trained model on PolyU fingerprint database.}
\label{fig:polyU_generate}
\end{figure*}

In Figure \ref{fig:fvc2006_generate}, we show the generated fingerprint images for four different latent representations, over every tenth epochs, when the model is trained on FVC2006 fingerprint dataset.
This will provide the granular changes happen after training the model for 10 more epochs.
As we can see, the generative model keeps improving over time, thus generating more realistic fingerprint images images.

In Figure \ref{fig:polyU_generate}, we show a similar results generating four samples images, using the trained model on PolyU fingerprint dataset.
We can make the same observation as above, where the generated sample images gets more realistic over time.

\begin{figure}[tbh]
\begin{center}
   \includegraphics[width=0.95\linewidth, trim={0 1cm 0 0}]{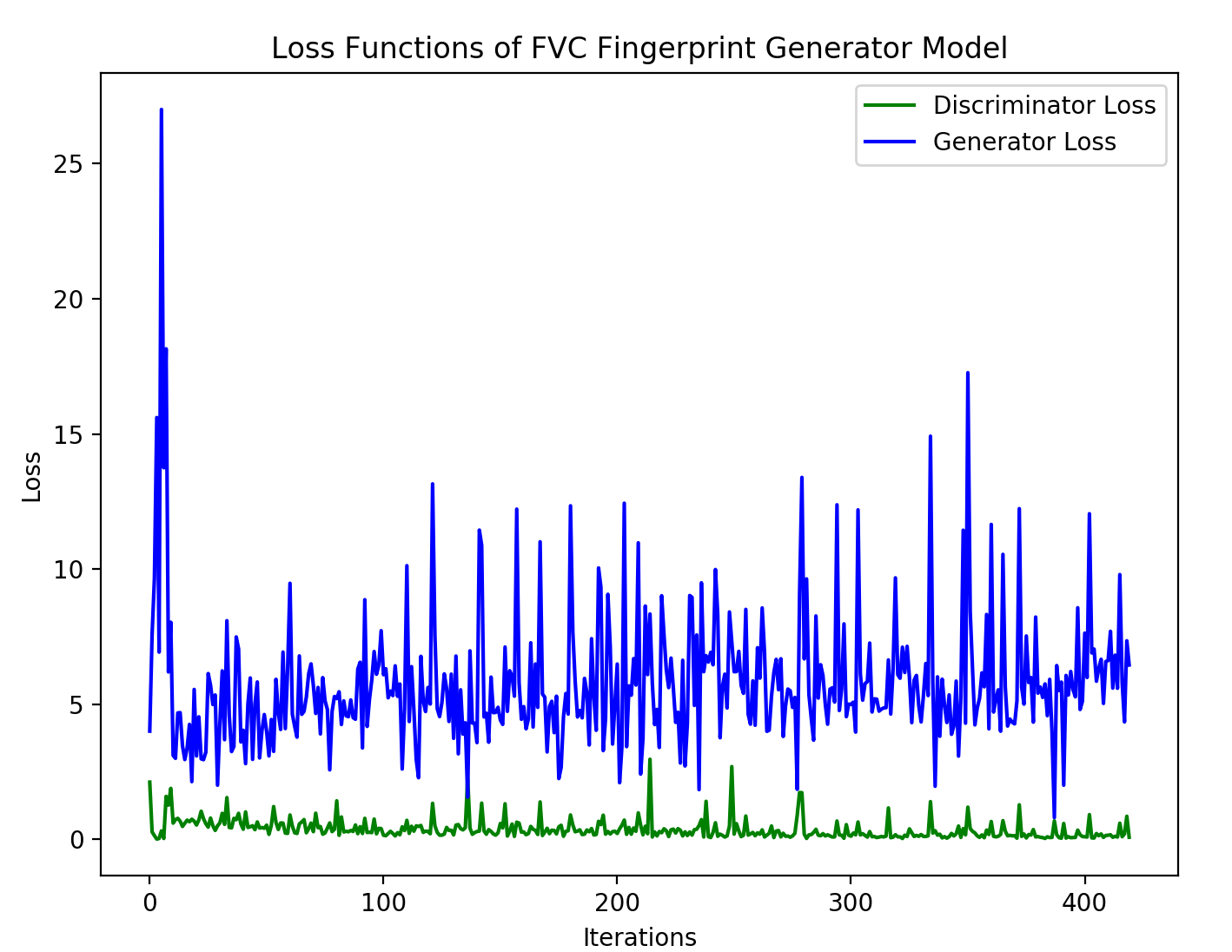}
\end{center}
   \caption{The discriminator and generator loss functions over different iterations, for the FVC-2006 fingerprint model.}
   \vspace{-6mm}
\label{fig:FVCloss}
\end{figure}

\begin{figure}[tbh]
\begin{center}
   \includegraphics[width=0.95\linewidth, trim={0 1cm 0 0}]{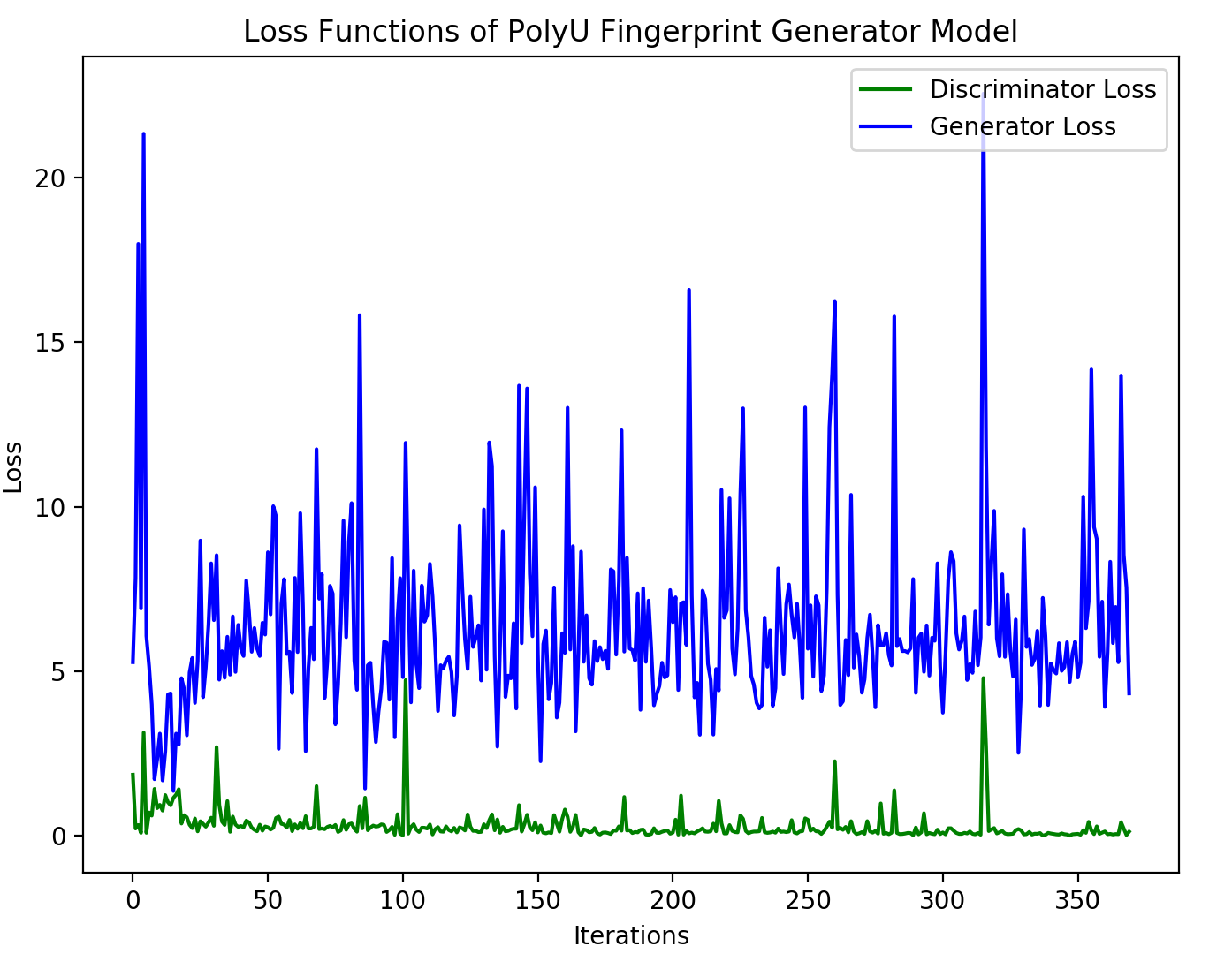}
\end{center}
\vspace{-2mm}
   \caption{The discriminator and generator loss functions over different iterations for the PolyU fingerprint model.}
   \vspace{-6mm}
\label{fig:polyU_loss}
\end{figure}

We also present the discriminator and generator loss functions for the trained Finger-GAN models on FVC2006 and PolyU databases.
The loss functions for these models are presented in Figures \ref{fig:FVCloss} and \ref{fig:polyU_loss}.

\subsection{Numerical Results}
To  measure  the quality  and  diversity of the model numerically,  we computed the Frechet Inception Distance (FID) \cite{FID} on the generated iris images by our model.
FID is an extension of Inception Score (IS) \cite{IS}, which was previously used for assessing the quality of generated images by GANs and other generative models.
FID compares the statistics of generated samples to real samples, using the Frechet distance between two multivariate Gaussians, defined as below:
\begin{equation}
\begin{aligned}
& \textbf{FID}=  \|\mu_r - \mu_g \|^2 + \textbf{Tr} (\Sigma_r + \Sigma_g - 2(\Sigma_r \Sigma_g)^{(1/2)} )
\end{aligned}
\end{equation}
where $X_r \sim N(\mu_r, \Sigma_r)$ and $X_g \sim N(\mu_g, \Sigma_g)$ are the 2048-dimensional activations of the Inception-v3 pool3 layer for real and generated samples respectively.

The FID scores by our proposed FingerGAN model is shown in Table \ref{TblComp}.
As we can see, the model achieves relatively low FID score, comparable with some of state-of-the-arts models on public image datasets.

\begin{table}[ht]
\centering
  \caption{The Frechet Inception Distance for the trained finger-gan model}
  \centering
\begin{tabular}{|m{4cm}|m{2.8cm}|}
\hline
Model/Database  & Frechet Inception Distance \\
\hline
DC-GAN/ FVC Fingerprint  &  \ \ \ \ \ \ \ \ \ \ \ \ \ 70.5 \\
\hline
\end{tabular}
\label{TblComp}
\end{table}

\section{Conclusion}
\label{sec:Conclusion}
In this work we propose a framework for fingerprint image generation using deep convolutional generative adversarial network. 
We use simple architectures for the generator and discriminator networks of our model, each containing 5 layers.
We also add a suitable regularization term  to our framework (the total variation of the output), to impose the connectivity of the generated fingerprint images.
These models are trained on two popular fingerprint databases, FVC 2006 and PolyU.
The experimental results show the synthetic fingerprint images look very realistic and similar to real fingerprints.
In the future, we plan to develop a model for fingerprint generation based on conditional-GAN model, which is able to generate image samples for a given identity.

\section*{Acknowledgment}
We would like to thank the Hong Kong Polytechnic University (PolyU) and FVC group for providing us the fingerprint databases used in this work.
We would also like to thank Ian Goodfellow for his comments and suggestions regarding the model performance analysis.

\end{document}